\documentclass{article}

\usepackage{microtype}
\usepackage{graphicx}
\usepackage{subfigure}
\usepackage{booktabs} 
\usepackage[table,xcdraw]{xcolor}
\usepackage{hyperref}

\usepackage[accepted]{icml2025}

\usepackage{amsmath}
\usepackage{amssymb}
\usepackage{mathtools}
\usepackage{amsthm}

\usepackage[capitalize,noabbrev]{cleveref}

\theoremstyle{plain}

\theoremstyle{definition}

\theoremstyle{remark}

\definecolor{table_green}{HTML}{B3EFB3}
\definecolor{table_blue}{HTML}{DAE8FC}
\usepackage[textsize=tiny]{todonotes}

\begin{document}

\twocolumn[
\icmltitle{Scaling Decentralized Learning with FLock}




\begin{icmlauthorlist}
\icmlauthor{Zehua Cheng}{oxford,flock}
\icmlauthor{Rui Sun}{nc}
\icmlauthor{Jiahao Sun}{flock}
\icmlauthor{Yike Guo}{hkust,hkgai}
\end{icmlauthorlist}

\icmlaffiliation{oxford}{Department of Computer Science, University of Oxford, Oxford, UK}
\icmlaffiliation{flock}{FLock.io, London, UK}
\icmlaffiliation{nc}{Newcastle University, UK}
\icmlaffiliation{hkust}{Hong Kong University of Science and Technology, Hong Kong SAR}
\icmlaffiliation{hkgai}{HongKong Generative AI
Research \& Development Center, Hong Kong SAR}

\icmlcorrespondingauthor{Zehua Cheng}{zehua.cheng@cs.ox.ac.uk}
\icmlcorrespondingauthor{Yike Guo}{yikeguo@ust.hk}

\icmlkeywords{Machine Learning}
\vskip 0.3in
]



\printAffiliationsAndNotice{}  

\begin{abstract}
Fine-tuning the large language models (LLMs) are prevented by the deficiency of centralized control and the massive computing and communication overhead on the decentralized schemes. While the typical standard federated learning (FL) supports data privacy, the central server requirement creates a single point of attack and vulnerability to poisoning attacks. Generalizing the result in this direction to 70B-parameter models in the heterogeneous, trustless environments has turned out to be a huge, yet unbroken bottleneck. 
This paper introduces FLock, a decentralized framework for secure and efficient collaborative LLM fine-tuning. Integrating a blockchain-based trust layer with economic incentives, FLock replaces the central aggregator with a secure, auditable protocol for cooperation among untrusted parties.
We present the first empirical validation of fine-tuning a 70B LLM in a secure, multi-domain, decentralized setting. Our experiments show the FLock framework defends against backdoor poisoning attacks that compromise standard FL optimizers and fosters synergistic knowledge transfer. The resulting models show a 68\% reduction in adversarial attack success rates. The global model also demonstrates superior cross-domain generalization, outperforming models trained in isolation on their own specialized data.

\end{abstract}

\section{Introduction}
The contemporary artificial intelligence (AI) landscape is characterized by centralization of power. The development of state-of-the-art foundation models, particularly Large Language Models (LLMs), necessitates vast quantities of data and extraordinary computational resources, concentrating this capability within a small number of large technology corporations. This concentration poses significant and systemic risks, including the potential for algorithmic bias amplification, a lack of transparent governance, heightened data privacy concerns for users, and the creation of single points of failure that could have widespread impact~\cite{dong2024defending}. The very architecture of modern AI development, predicated on massive, centralized data aggregation, is in direct tension with the growing global imperative for data sovereignty and individual privacy.

Federated Learning (FL) has emerged as the leading technical paradigm to resolve this tension. At its core, FL allows multiple parties to collaboratively train a shared machine learning model without exchanging or centralizing their raw, private data. In the canonical approach, Federated Averaging (FedAvg)~\cite{mcmahan2017communication}, clients locally train a model on their respective datasets and submit only the resulting model updates (e.g., gradients or weights) to a central server, which then aggregates them to produce an improved global model~\cite{geiping2020inverting}. This process preserves data locality, offering a powerful alternative to traditional centralized training.

But the vanilla FL model is not a silver bullet. Its use of a central server for aggregation brings back a kind of centralization, with a key single point of failure and an attractive value proposition for attackers~\cite{kairouz2021advances,li2020federated}. If this central aggregator is compromised, the entire learning process can be controlled, against the wishes of all participants~\cite{konevcny2016federated}. Even more fundamentally, the FL model inevitably tends to work on a naive assumption of honest clients~\cite{bhagoji2019analyzing}. This is an unrealistic proposition in open, decentralized environments. Malicious clients may mount advanced poisoning attacks, either by injecting corrupted data into their local training sets (data poisoning) or by hacking their submitted model updates (model poisoning), with an aim to disrupt the global model's accuracy or to inject a focused backdoor. 
This research field must therefore advance beyond merely maintaining privacy to building verifiable trust and robustness, within a decentralized environment.

This paper presents the FLock~\cite{dong2024defending} system as a comprehensive solution to these intertwined challenges, specifically in the demanding context of fine-tuning massive LLMs. Our work is built upon the thesis that by integrating a blockchain-based integrity layer with a novel, communication-efficient optimization strategy, it is possible to achieve what was previously considered impractical: the secure, scalable, and synergistic fine-tuning of 70B-parameter LLMs across heterogeneous, multi-domain clients. The research detailed herein makes three principal contributions to the state of the art:

\begin{itemize}
    \item We provide a conclusive, empirical validation that shatters the long-standing scalability barrier in decentralized learning. It irrefutably demonstrates the successful fine-tuning of a 70B LLM in a secure, multi-domain setting, a feat previously deemed impossible in such contexts.
    \item We present a FL with Blockchain system, FLock, that unifies a blockchain-based trust layer with communication-efficient optimization to concurrently address security, scalability, and efficiency.
    \item Our experimental results demonstrates that the FL with Blockchain yields statistically significant performance improvements over isolated, single-client training. The collaboratively trained models exhibit substantially enhanced adversarial robustness, reducing the Attack Success Rate (ASR) by more than 68\% compared to the baseline model. Furthermore, the work reports a synergistic knowledge transfer, where the resulting global model shows superior cross-domain generalization and can outperform models trained exclusively on a single domain's data, even on their own specialized test sets.
\end{itemize}

\section{Related Works}
\subsection{Federated Learning}
Federated Learning (FL) trains a shared model across multiple nodes holding local data samples and coordinated by a central node that aggregates their individual model updates.
In each communication round, a central server distributes the current global model to a subset of clients. 
These clients then perform multiple local training steps (e.g., using SGD) on their private data. 
The resulting updated local models are sent back to the server, which aggregates them, typically through a weighted average based on the size of each client's dataset, to produce the next iteration of the global model. This cycle repeats until the model converges. As FL is designed for a a privacy-preserving model training solution~\cite{mcmahan2017communication}, recent exploration on the attack on the centralized server has raised the concern of the safety of centralized FL solution~\cite{zhuang2023foundation}. Therefore, decentralized FL~\cite{xing2021federated} has become an optimal option to address such concerns.

While elegant in its simplicity, this process faces significant hurdles in real-world deployments~\cite{zhuang2023foundation}. 
Most prominently among these is the non-independent and identically distributed (non-IID) property of client data. In practice, data distribution varies from client to client due to patterns in users, geography, or any other factor. This diversity has the power to cause optimization paths of local models to be much different from the global optimum, something we call ``client drift''. Client drift has the ability to severely disrupt the convergence of the global model and lead to bad performance and instability. This issue specifically induces our experimental approach, which openly uses a multi-domain scenario to push the system to accommodate extreme data diversity.

\begin{figure*}[t]
    \centering
    \includegraphics[width=\linewidth]{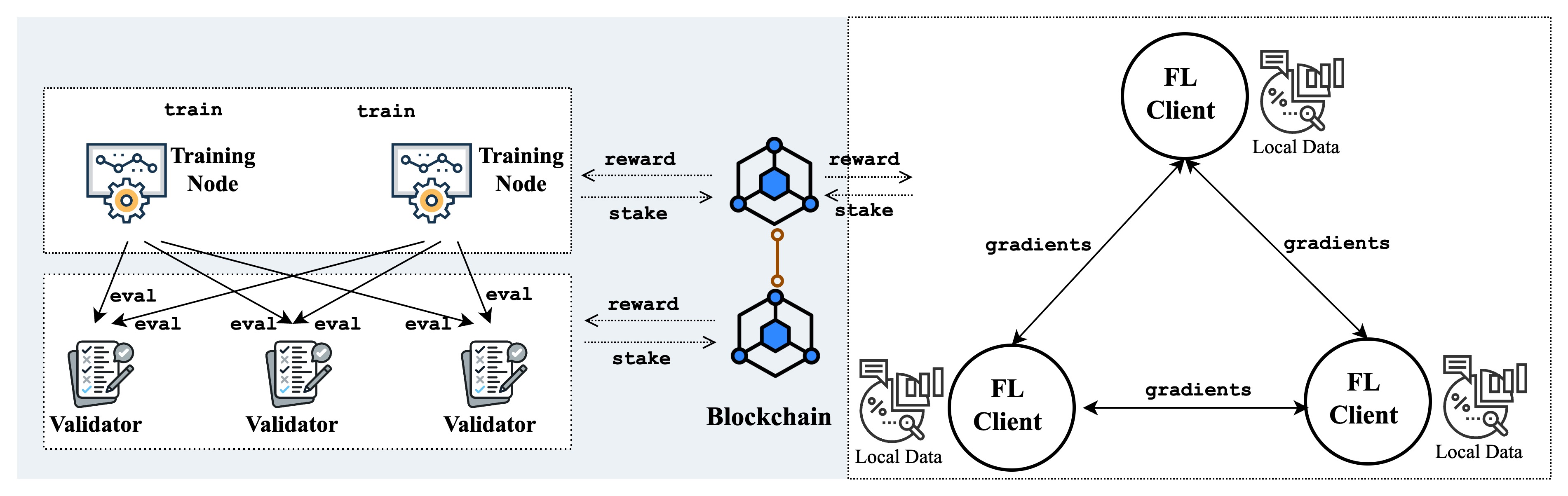}
    \caption{General Design of FLock Network for Decentralized Large-language Model Finetuning.}
    \label{fig:overall}
\end{figure*}

\subsection{Securing Federated Learning with Blockchain}
The inherent trust deficit in centralized FL, where there is one server to coordinate and aggregate on, has prompted researchers to consider decentralized equivalents. As a technological approach, blockchain, by virtue of its very characteristics of immutability, transparency, and distributed consensus, offers a very attractive architectural solution. By recording model updates and aggregation results to an immutable record, a blockchain-based system offers a provable and auditible record of the entire training procedure which renders unnecessary placing trust in some central agent.

This line of research is foundational to our work. Prior publications, such as FLock~\cite{dong2024defending} have established a robust framework for secure, decentralized FL. The core innovation of these systems is the replacement of centralized trust with a combination of cryptographic verification and economic incentives. Smart contracts, which are self-executing programs on the blockchain, are used to automate the model aggregation process transparently. To defend against malicious clients who might submit poisoned updates, these systems introduce a peer-to-peer (P2P) review mechanism, where a committee of participants (``voters'') evaluates the quality of aggregated updates. This is coupled with a reward-and-slash mechanism, where participants stake collateral. Honest contributions are rewarded from this pool, while malicious or low-quality contributions are penalized (``slashed''), creating a strong economic disincentive for attacks. Our current work leverages this established security architecture as the trusted foundation upon which we build our scalable LLM fine-tuning protocol.

\subsection{Scaling LLM in Decentralized Setup}
Applying any form of FL to LLMs, particularly ones running 70 billion, borders on the impractical. The fundamental bottleneck is the sheer size of the model updates alone. The full 32-bit gradient update to a 70B parameter model would require sending about 280 GB of data per client, per iteration. Alongside the sheer local RAM required to maintain the model and its gradients, end-to-end fine-tuning becomes impractical on anything besides the best-funded institutional clientele, making a decentralized ecosystem impossible. This ``scalability wall'' constitutes the fundamental problem that our work aims to solve.

Several lines of research have emerged to tackle this challenge. Parameter-Efficient Fine-Tuning (PEFT) techniques, such as Low-Rank Adaptation (LoRA)~\cite{hu2022lora}, have become standard practice. PEFT methods freeze the vast majority of the pre-trained model's weights and introduce a small number of new, trainable parameters (adapters). This dramatically reduces the computational and memory footprint of fine-tuning. However, even when applying PEFT in a federated setting, the communication costs can remain substantial, and naive application can lead to performance degradation compared to centralized training.

Other research has focused on building systems for distributed LLM execution. Systems like Petals enable collaborative inference and fine-tuning of massive models over the internet by distributing layers of the model across different consumer-grade devices. While these efforts demonstrate the feasibility of running large models decentrally, their primary focus is often on model parallelism and fault tolerance rather than the specific threat model of adversarial poisoning attacks that FLock is designed to mitigate. Frameworks like FedPepTAO~\cite{CheL00ZSDD23} have also been proposed to make prompt tuning more communication-efficient in FL, highlighting the community's focus on reducing the communication burden.

The literature reveals a crucial void. There exist FL security solutions (like FLock), LLM efficiency solutions (like PEFT), and communication overhead solutions (like gradient compression). However, those are not really put together inside one, end-to-end, system that's secure, highly scalable to 70B models, and yet still efficient to fit inside a heterogeneous network. That's new work precisely because such an end-to-end system is put forth and demonstrably exists, and yet meets those three requirements at once: security, scale, and efficiency.

\section{Methodologies}
The FLock system fundamentally re-architects federated learning to eliminate the reliance on a central, trusted server. It achieves this by leveraging distributed ledger technology (DLT) and a carefully designed set of economic incentives managed by on-chain smart contracts. This architecture, detailed in foundational work, provides the integrity and security necessary for collaboration among untrusted participants. The system operates through defined roles and distinct phases. 

\subsection{Decentralized tuning with FLock}
We present the overall framework of FLock network for decentralized LLM fine-tuning in Figure~\ref{fig:overall}. The system architecture comprises two principal entities: Training Node and Validator.

\textbf{Training Node} are in charge of carrying out the process of training the model in a given task using shared data. To be engaged in this operation, one has to stake blockchain assets or tokens, making eligibility staking-dependent. Rewards of the Training Node are paid through a weighted combination of their staked collateral alongside their perceived performance. 

\textbf{Validator} are in charge of the validation of work carried out by the Training Nodes. These present validation scores that are used directly in the calculation of reward distribution. Like training nodes, validators are also required to stake tokens in order to be active. The staking has a twin function: they are given the rights to validate tasks that are assigned to them, and they are ensured fair distribution of validation work. 

To setup the training node, the local dataset denoted as $\mathcal{D}_{\mathrm{local}}$ consists of locally sourced data samples, which are composed of a feature set $X$ and a corresponding label set $Y$. For each feature vector $x_i \in X$, there is an associated true label $y_i \in Y$. The objective is to train a predictive model, $f$, which learns the underlying patterns within $\mathcal{D}_{\mathrm{local}}$ to accurately map features to their corresponding labels, such that $f(x_i) \approx y_i$.

\subsection{Reward Distribution Protocol}

This section details the entire process of distribution of the reward of a task. It will motivate high-quality submissions of the model by the training nodes as well as accurate judgments by validators. It considers an environment where there are $n$ training nodes, providing model submissions $O_i$ with stake $t_i$, and $m$ validators, providing a stake $s_j$. 

\subsubsection{Stake-Weighted Consensus Score}

Before rewarding, a final performance score of all submitted models is derived by taking a stake-weighted combination of all validator judgments. Each validator $V_j$ produces a score vector $\vec{r}_j = (r_{j1}, \ldots, r_{jn})$ of the $n$ submissions. The final combined score vector, $\vec{r}$, can be computed as follows:

$$\vec{r} = \left( \frac{\sum_{j=1}^{m} r_{j1} \cdot s_j}{\sum_{j=1}^{m} s_j}, \ldots, \frac{\sum_{j=1}^{m} r_{jn} \cdot s_j}{\sum_{j=1}^{m} s_j} \right)$$

This process ensures that evaluations from validators with higher stakes exert greater influence on the consensus outcome, thereby leveraging economic incentives to foster reliable assessments.

\subsubsection{Task-Level Reward Allocation}

In a given task, the day-long reward, $R_0$, at the beginning, is split between the validators group and the training nodes group. It is split in accordance with their corresponding total stakes. $\gamma$ controls this ratio, inclining the balance towards a constant reward component, as opposed to a stake-dependent component. 
The total reward pool $R$ for all training nodes is defined as: 
\begin{equation}
    R_{\text{train}} = R_0 \cdot \left( \gamma + (1-2\gamma) \cdot \frac{\sum_{i=1}^{n} t_i}{\sum_{i=1}^{n} t_i + \sum_{j=1}^{m} s_j} \right)
\end{equation}
Similarly, the total reward pool for all validators:
\begin{equation}
    R_{\text{val}} = R_0 \cdot \left( \gamma + (1-2\gamma) \cdot \frac{\sum_{j=1}^{m} s_j}{\sum_{i=1}^{n} t_i + \sum_{j=1}^{m} s_j} \right)
\end{equation}

\subsubsection{Individual Rewards for Training Nodes}
The reward of a training node depends upon its rank of performance as well as its total staking amount. 
\begin{itemize}
    \item \textbf{Performance Ranking}: Ranks are given to training nodes between $k=1$ and $n$, based on the cumulative $\vec{r}$ values. Rank-weight, in the geometric series form, $g_k$, is calculated such that larger-rank nodes are weighted disproportionately:$g_k = \frac{1-q}{1-q^n} \cdot q^{k-1}$. The rank of the node, the number of training nodes, are represented by $k$, respectively, while $q \in (0,1)$ represents the common ratio of the series.
    \item \textbf{Reward Share Calculation}: The proportion of $R_{\text{trainers}}$ that training node $i$ is entitled to, denoted as $\text{Share}_i$, is calculated by combining its rank-based weight $g_i$ with its total stake $t_i$: $\text{Share}_i = \frac{g_i \cdot t_i^{\alpha_t}}{\sum_{k=1}^{n} g_k \cdot t_k^{\alpha_t}}$. The parameter $\alpha_t$ controls the influence of the stake amount on the final reward. 
    \item \textbf{Delegation Split}: If the total stake $t_i$ of a training node consists of its own stake $t_{\text{node}}$ plus a delegated stake $t_{\text{delegate}}$ (i.e., $t_i = t_{\text{node}} + t_{\text{delegate}}$), the reward that the node operator earns is influenced by a commission rate $\sigma$.
\end{itemize}

\begin{equation}
    \mathcal{R}_{\text{node}, i} = f_i(g_i, t_i) \cdot \left( \sigma + (1-\sigma) \cdot \frac{t_{\text{node}}}{t_{\text{node}} + t_{\text{delegate}}} \right)
\end{equation}

\subsubsection{Validator Rewards for Individuals}

Validators receive their reward in direct proportion to the accuracy of their judgments, that being the closeness of their scores to the eventual stake-weighted consensus. 

\begin{itemize}
    \item \textbf{Accuracy Measurement}: For each validator $V_j$, 
    we compute a distance vector $\vec{\Delta}_j$ to quantify the deviation of its scores from the consensus score $\vec{r}$, which is defined in Equation~\ref{eq:acc_meassurement}, where $r_i$ is the $i$-th component of the aggregated score vector $\vec{r}$.
    \item \textbf{Reward Share Calculation}: A distribution function, based on a modified Softmax, determines a validator's reward share for its evaluation of each model $i$. This function, $f_i(\Delta_{ji}, s_j)$, is designed to decrease with greater distance $\Delta_{ji}$ and increase with a larger stake $s_j$, which is defined in Equation~\ref{eq:reward_share_cal} where $\lambda_v$ controls the sensitivity to evaluation accuracy, while $\alpha_v$ determines the influence of the stake amount.
    \item \textbf{Total Reward and Delegation}: Validator $V_j$'s overall reward share in all $n$ validations is the sum of its individual shares, $\sum_{i=1}^{n} f_i(\Delta_{ji}, s_j)$. Its share is then applied to the validator reward pool $R_{\text{validators}}$. Similarly for training nodes, in the case the validator's stake $s_j$ has delegated funds (i.e., $s_j = s_{\text{validator}} + s_{\text{delegated}}$), the operator's final reward is pro-rated by its configured commission rate $\sigma$, 
\end{itemize}

\begin{equation}\label{eq:acc_meassurement}
    \vec{\Delta}_j = \left( \left| r_1 - r_{j1} \right|, \ldots, \left| r_n - r_{jn} \right| \right)
\end{equation}

\begin{equation}\label{eq:reward_share_cal}
    f_i(\Delta_{ji}, s_j) = \frac{e^{-\lambda_v \Delta_{ji}} \cdot s_j^{\alpha_v}}{\sum_{l=1}^{m} e^{-\lambda_v \Delta_{li}} \cdot s_l^{\alpha_v}}
\end{equation}

\begin{equation}
\label{eq:total_reward}\small
    \mathcal{R}_{\text{val}, j} = (R_{\text{val}} \cdot \sum_{i=1}^{n} f_i(\Delta_{ji}, s_j)) \cdot \left( \sigma + (1-\sigma) \cdot \frac{s_{\text{val}}}{s_{\text{val}} + s_{\text{delegated}}} \right)
\end{equation}

\section{Experimental Results}

\subsection{Experimental Setup}

This setup mimics a practical, real-world federated learning (FL) environment of diverse, domain-specialized agencies. Our goal is to fine-tune a Qwen2.5 72B general-purpose large language model~\cite{qwen25} in a consortium of eight simulated government agencies, in turn, each of which we represent by a separate, high-quality dataset. Here, we describe the hardware setup, the choice of datasets, and the reasons behind the experimental setup, in its ability to replicate the real-world, multi-stakeholder AI collaboration problems.

These experiments ran on a high-end computing cluster of eight server nodes. Each of the servers has eight NVIDIA H800 GPUs, connected by high-speed NVLink. For orchestrating the process of distributed training and attaining the best possible performance through this 64-GPU system, we utilized the DeepSpeed library~\cite{rasley2020deepspeed}, the latest-of-breed of large-scale model training frameworks. Such powerful infrastructure allows us to realistically test the system's performance under heavy computational workloads. 

One of the greatest challenges in federated learning is the statistical heterogeneity, in that the data distribution of the clients involved in the system is non-independent and identically distributed (non-IID). Each of the data sets is a separate "client" in our federated system, akin to a separate government bureau or department that has specialized data. The non-IID environment plays a crucial role in the global model being able to learn from and generalize across diverse data sources without sacrificing domain-specificity. 
\begin{table*}[t]\centering\small
\begin{tabular}{c|cccc|c|cc|cc}\toprule
Node           & MT1           & MT2           & MT-Bench & GPT-4 Judge   & Average       & AdvBench ($\downarrow$) & AB-GCG ($\downarrow$) & HumanEval     & MMLU          \\\toprule
Baseline       & 9.53          & 8.92          & 9.35     & -             & -             & 9.3                                  & 38.3                                     & 86.6          & 85.8          \\\midrule
Local Training                    & 9.18          & 8.85          & 9.07     & 8.64          & 8.94          & 14.5                    & 45.1                  & 84.3          & 86.5          \\ \hline
\rowcolor[HTML]{B3EFB3} Client\#1 & 9.25          & 8.9           & 9.15     & 8.70          & 9.00          & 15.1                    & 46.2                  & 83.1          & 86.9          \\
\rowcolor[HTML]{B3EFB3} Client\#2 & 9.15          & 8.81          & 9.05     & 8.65          & 8.91          & 14.8                    & 44.8                  & 84.5          & 86.3          \\
\rowcolor[HTML]{B3EFB3} Client\#3 & 9.30           & 9.01          & 9.22     & 8.80          & 9.08          & 13.9                    & 42.5                  & 85.2          & 86.2          \\
\rowcolor[HTML]{B3EFB3} Client\#4 & 9.05          & 8.70           & 8.90     & 8.51          & 8.79          & 16.2                    & 48.9                  & 82.0            & 85.1            \\
\rowcolor[HTML]{B3EFB3} Client\#5 & 9.01          & 8.65          & 8.85     & 8.45          & 8.74          & 15.8                    & 47.1                  & 81.5          & 84.8         \\
\rowcolor[HTML]{B3EFB3} Client\#6 & 9.22          & 8.91          & 9.13     & 8.72          & 8.99          & 14.1                    & 43.4                  & 86.0            & 86.6           \\
\rowcolor[HTML]{B3EFB3} Client\#7 & 9.20           & 8.88          & 9.10     & 8.68          & 8.96          & 15.0                      & 45.5                  & 83.9          & 86.9          \\
\rowcolor[HTML]{B3EFB3} Client\#8 & 9.10           & 8.95          & 9.08     & 8.60          & 8.93          & 13.5                    & 42.0                    & 88.1          & 86.3          \\ \midrule
\rowcolor[HTML]{DAE8FC} 
Client\#1      & 9.77          & 9.11          & 9.50     & 8.80          & 9.29          & 6.3                                  & 12.5                                     & \textbf{88.3} & 85.5          \\
\rowcolor[HTML]{DAE8FC} 
Client\#2      & 9.78          & 9.11          & 9.44     & 8.50          & 9.20          & 7.5                                  & 12.3                                     & 87.8          & 85.8          \\
\rowcolor[HTML]{DAE8FC} 
Client\#3      & 9.69          & 9.05          & 9.43     & 8.90          & 9.26          & \textbf{7.3}                         & 19.5                                     & 87.8          & 85.7          \\
\rowcolor[HTML]{DAE8FC} 
Client\#4      & 9.55          & 9.06          & 9.53     & 8.81          & 9.23          & \textbf{7.3}                         & 13.3                                     & 87.5          & 86.1          \\
\rowcolor[HTML]{DAE8FC} 
Client\#5      & \textbf{9.87} & 9.11          & 9.53     & 8.90          & \textbf{9.35} & 7.5                                  & \textbf{12.1}                            & 86.6          & 86.8          \\
\rowcolor[HTML]{DAE8FC} 
Client\#6      & 9.63          & 9.10          & 9.44     & 8.88          & 9.26          & 7.4                                  & 19.5                                     & 87.3          & 86.3          \\
\rowcolor[HTML]{DAE8FC} 
Client\#7      & 9.70          & \textbf{9.20} & 9.53     & \textbf{8.93} & 9.34          & 7.5                                  & \textbf{12.1}                            & 86.5          & 86.5          \\
\rowcolor[HTML]{DAE8FC} 
Client\#8      & 9.78          & 9.11          & 9.44     & 8.80          & 9.28          & 8.5                                  & 12.5                                     & 87.5          & \textbf{86.8}\\\bottomrule
\end{tabular}\label{tab:overall}
\caption{Experimental Results on Open-ended and Closed-ended Evaluations with FLock System. The experiments presented in \colorbox{table_green}{local-only} training and \colorbox{table_blue}{FL} training.
For MT-Bench, the MT-1 is aim to evaluate responses to single-turn questions. For AdvBench, we reported the Average Success Rate (ASR) which is the lower the better. AB-GCG is applying the Greedy Coordinate Gradient (GCG)~\cite{zou2023universal} and evaluated in AdvBench. For all metrics except AdvBench and AB-GCG, higher scores indicate better performance. For the adversarial benchmarks, lower scores are better, signifying greater model robustness.}
\end{table*}

\begin{itemize}
    \item \textbf{Legal Domain} (Client\#1): LegalBench~\cite{guha2023legalbench}: This benchmark consists of 162 tasks designed to evaluate complex legal reasoning. We utilized a subset of these tasks focused on statutory interpretation and compliance, reflecting the analytical needs of a governmental legal department. The inherent complexity and specialized vocabulary of LegalBench allow us to test the model's capacity for nuanced, domain-specific text comprehension within the federated setting.
    \item \textbf{Medical Domain} (Client\#2): PubMedQA~\cite{jin2019pubmedqa}: A biomedical question-answering dataset comprising over 273,000 instances derived from PubMed abstracts. The task requires a yes/no/maybe response to research questions based on scientific text. This dataset simulates the data environment of a public health agency, where evidence-based reasoning is critical and data privacy is paramount.
    \item \textbf{Education Domain} (Client\#3): AI2~\cite{ai2challenge2024}: The ARC dataset contains 7,787 multiple-choice science questions from grade-school examinations. These questions often require multi-step reasoning and a foundational understanding of scientific concepts, representing the challenges faced by an educational agency in curriculum development and assessment.
    \item \textbf{Environmental Impact \& Planning} (Client\#4): NEPATEC1.0~\cite{PolicyAI2024NEPATEC}: A large-scale corpus of over 28,000 documents (4.8 million pages) from the National Environmental Policy Act (NEPA) database. This dataset, rich with technical and regulatory language, is representative of the data utilized by an environmental protection agency for impact assessment and policy enforcement.
    \item \textbf{Civic Discourse} (Client\#5): MeetingBank~\cite{huetal2023meetingbank}: This dataset includes 1,366 transcripts from U.S. city council meetings. Its long-form, conversational nature provides a challenging test for summarization and topic extraction, tasks central to the operations of public-facing government bodies that must document and analyze civic engagement.
    \item \textbf{Finance Domain} (Client\#6): Finance Alpaca~\cite{gbharti2023financealpaca}: provides approximately 68,900 instruction-following examples tailored to the financial domain. It simulates the data a treasury or finance department would use, enabling the model to handle queries related to economic concepts, financial advice, and market analysis.
    \item \textbf{Regulatory Compliance} (Client\#7): PolicyBench~\cite{foo2025know}: This benchmark contains question-answer pairs based on public-facing government regulations and policies. By training on these examples, the model learns to interpret and answer user queries about specific rules and official guidelines, mimicking the function of a regulatory affairs or public-information office.
    \item \textbf{Finance Compliance} (Client\#8): EDGAR-CORPUS~\cite{loukas2021edgar} is built from over 90,000 corporate annual reports (10-K filings) filed between 1993 and 2020. Representing a financial oversight body, this dataset's billions of tokens provide a deep grounding in corporate finance, risk disclosure, and regulatory compliance language.
\end{itemize}

This curated combination of datasets creates a realistic proxy for the challenges of inter-agency collaboration. Governments do not operate on a single type of data; they require expertise spanning law, health, finance, environmental science, and public policy. Our experimental design directly addresses this reality by using datasets that are not only large and topically diverse but also varied in their format and task—from structured Q\&A to dense, unstructured reports. This heterogeneity prevents the model from overfitting to a single domain and forces the FLock system to synergistically merge knowledge from disparate sources. Successfully training a model in this environment serves as a powerful validation of its potential for real-world deployment in complex, multi-stakeholder government settings.

\subsection{Evaluation}
To assess the efficacy of our proposed training methodologies on realistic FL datasets, we consider 6 evaluation metrics, including 4 open-ended metrics and 2 closed-ended metrics.

\textbf{Open-ended evaluation}: Conversational proficiency, encompassing both single and multi-turn interactions, is measured using MT-Bench~\cite{zheng2023judging}. To assess model safety and alignment, we utilize AdvBench~\cite{zou2023universal}, which measures the rate of safe responses to adversarial prompts. We also present the GCG in~\cite{zou2023universal} to perform an attack based on AdvBench to show the robustness of our method. We report the Attack Success Rate (ASR) in the table. Complementing these, we introduce a bespoke in-domain metric, denoted as GPT-4 Judge. This involves scoring responses to 50 randomly sampled, unseen test samples. A GPT-4 Judge assesses the quality of the generated response against the ground-truth reference~\cite{ustun2024aya}, following the prompt template detailed in Appendix~\ref{appendix:llm-as-judge}.

\textbf{Close-ended evaluation}: we aim to verify that the fine-tuning process does not degrade foundational capabilities acquired during pre-training. To this end, we utilize two standard benchmarks. First, the Massive Multitask Language Understanding (MMLU) benchmark~\cite{dan2021MMLU} is used to measure the model's retained knowledge across 57 diverse subjects. Second, we employ HumanEval~\cite{chen2021evaluating} to assess the model's proficiency in generating executable code from docstrings. These metrics ensure that improvements in specialized tasks do not come at the cost of core, pre-trained competencies.

\subsection{Results on Benchmarks}
The experimental results, as presented in Table~\ref{tab:overall}, show a notable disparity between the performances of local training and the FL approach. The "Local Training" models represent randomly sampled two clients to run local training and average their evaluation results. The results show only marginal improvements over the "Baseline" Qwen2.5 72B model.

\begin{figure}[t]
    \centering
    \includegraphics[width=0.5\textwidth]{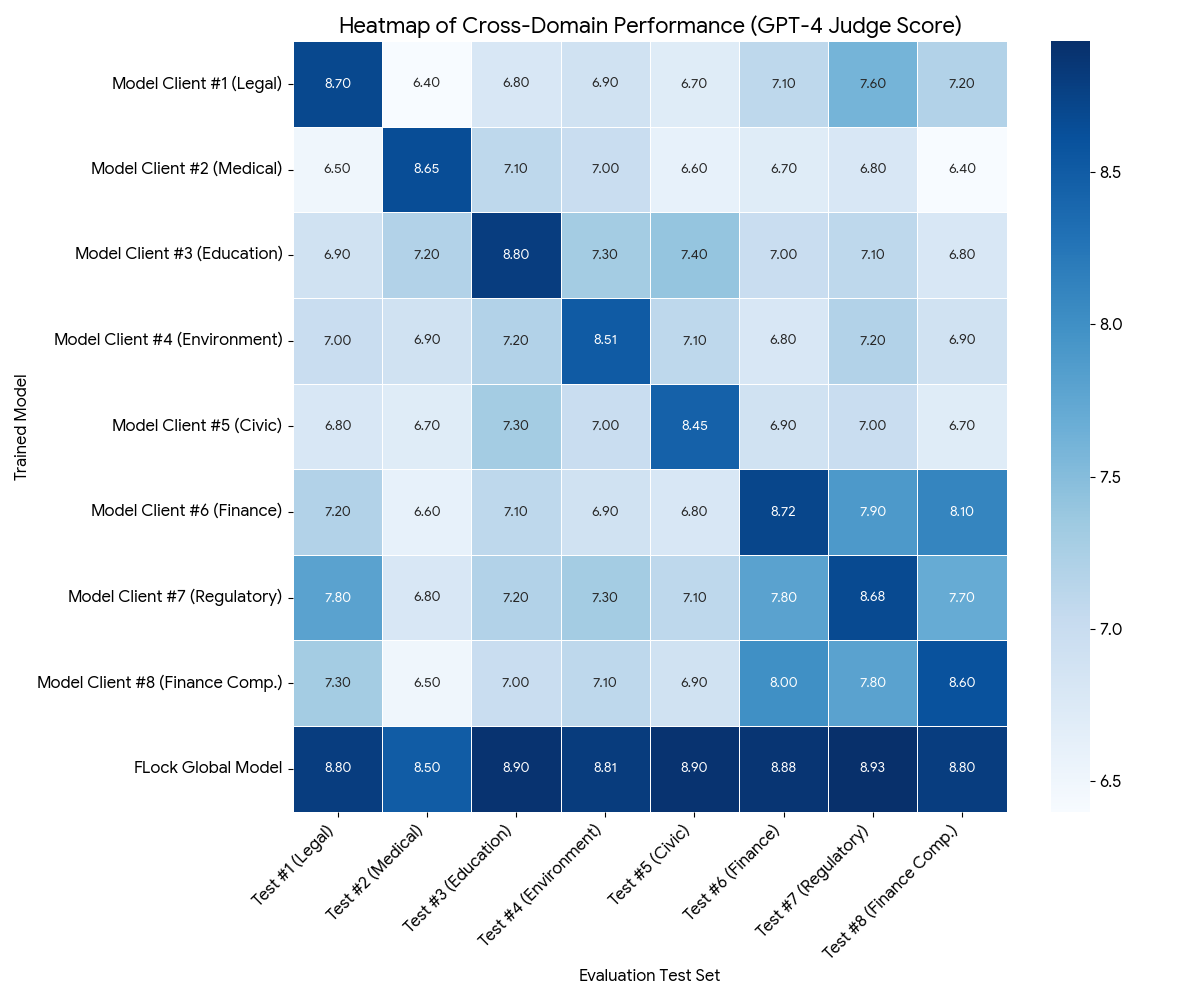}
    \caption{Heatmap of Cross-Domain Performance on the GPT-4 Judge Metric}
    \label{fig:heatmap}
\end{figure}

The results of \colorbox{table_green}{local-only} training, where the model was locally fine-tuned across individual-client datasets, reveals the significant limitations of this approach. While marginal and inconsistent variations in the performance were registered across general capability tests, model robustness suffered systematic erosion. 
For every client, the AdvBench and AB-GCG scores increased substantially compared to the baseline (e.g., rising to 16.2 and 48.9, respectively, for Client\#4). This consistent increase in scores indicates that specialization on a narrow, isolated dataset renders the model more brittle and significantly more vulnerable to adversarial attacks. This finding demonstrates that naive local fine-tuning is an inadequate, and even detrimental, strategy for specialized model adaptation.


Most notably, the \colorbox{table_blue}{FL} training achieves a substantial impovement in adversarial robustness. It reduces the AdvBench score from the baseline of 9.3 to as low as 6.3 (Client\#1). An even more considerable outcome is the one for the benchmark of the AB-GCG, where the FL models lower the score from the baseline 38.3 to at least 12.1 (Clients \#5 and \#7). This corresponds to over 68\% lower vulnerability from the baseline.

Significantly, these defense improvements accompany consistent improvements in general performance. The FLock models consistently surpassed the baseline by a considerable margin on exercises in reasoning and knowledge, with HumanEval scores at 88.3 (Client\#1) and MMLU scores at 86.8 (Client\#8) highs. Correspondingly in open questions by GPT-4, the score reached 8.93 (Client\#7), and in these the model's conversational and following-instruction strengths appear more pronounced. The federated learning framework used in the FLock system is an effective technique, producing models that not only generalize better at general-purpose tasks but also turn out to be more resilient to the impact of malicious threats in the guise of adversaries.

\subsection{Results on Cross Domain Generalization}

Figure~\ref{fig:heatmap} visualizes the performance of eight models trained in isolation (Rows 1-8) and the single collaborative FLock model (Row 9) across eight specialized test datasets. Darker shades of blue indicate higher performance scores. The visualization clearly contrasts the brittle, domain-specific knowledge of isolated models with the robust, generalized performance of the model trained using the FLock framework.
\begin{figure*}[t]
    \centering
    \includegraphics[width=0.9\textwidth]{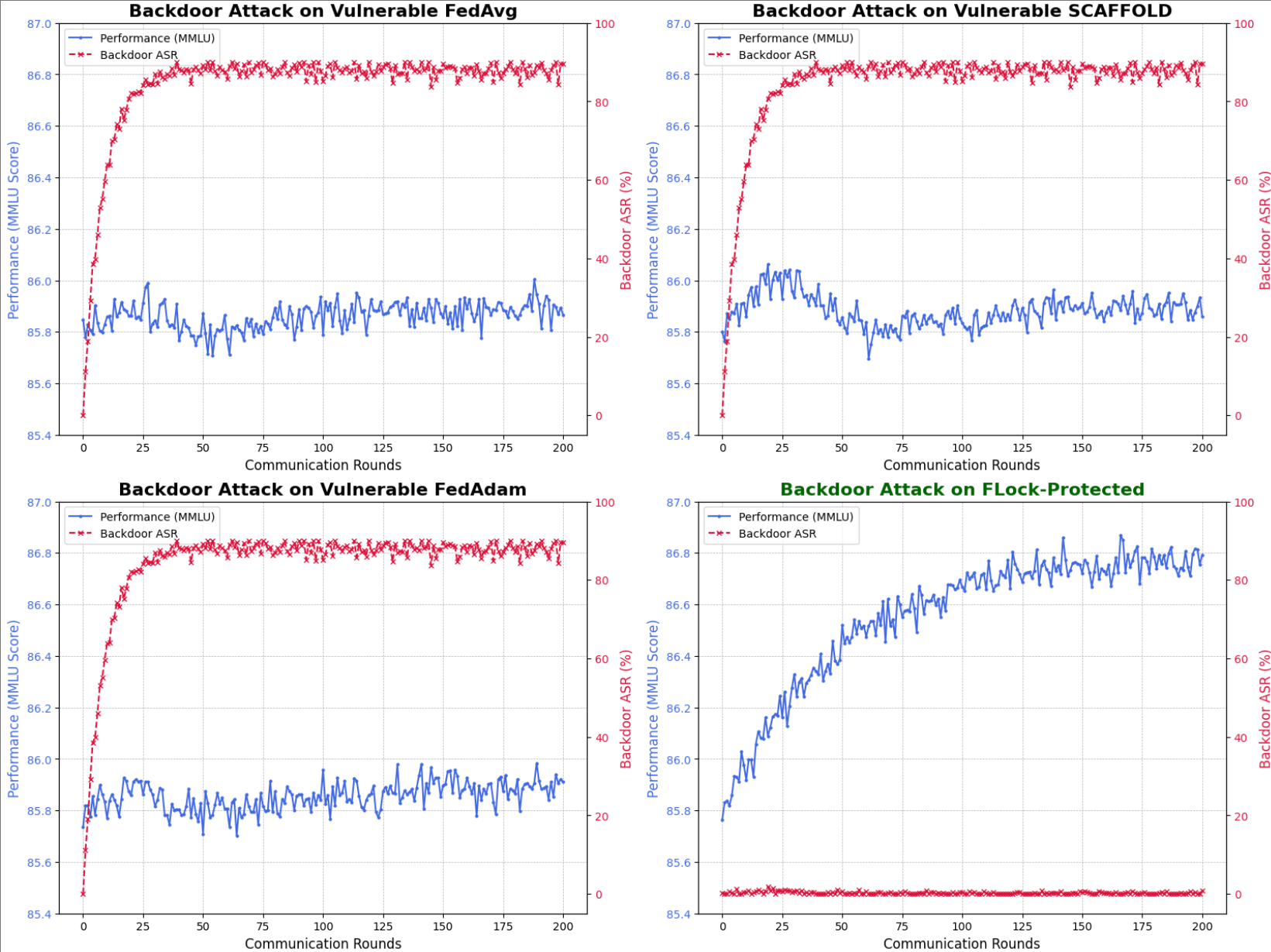}
    \caption{A comparative analysis of model performance (MMLU Score) and security (Backdoor ASR) under a sustained poisoning attack.}
    \label{fig:attack}
\end{figure*}
The dark diagonal line that runs from top to bottom through the top eight rows shows that each locally trained model excels only within its own special domain, while the off-diagonal white squares instantly call out the fundamental failure of this strategy: a sharp drop-off in performance when a model has to be tested on an unseen task. This behavior is a strong visual indication of the "brittleness" that comes from local training within data silos.
The bottom row, which shows the FLock Global Model, stands out from the rest. It's a solid, dark bar, indicating that this model consistently did very well on every one of the eight special-purpose test tasks. This indicates that the overall training procedure properly produces a strong, generalized model that isn't tied to the narrow specializations of any one participant's data.

The results presented a clear synergistic improvement on FLock framework. The most compelling evidence comes from comparing the specialists' performance to the performance of the FLock model. On many instances, the cell of the FLock model's cell is darker than that on the diagonal corresponding to the matching specialist. For example, the test-set individually trained finance and legal models are outperformed by the FLock model itself. This shows that FLock and not just average information but create a synergistic output, where knowledge transfer from one domain to another results in a model that has better performance than its constituents.

\section{Results on Resilience to Poisoning Attacks}
To empirically validate the security architecture of the FLock system, we designed a stress test to evaluate its resilience against a sophisticated backdoor poisoning attack. The objective was to determine if FLock's on-chain validation and economic incentive mechanisms could successfully defend the global model where standard federated learning protocols fail.

In this experiment, we designated a single participant, Client \#1 (Legal), as a malicious actor. The attacker's goal was not to degrade the model's general performance in a way that would be obvious on broad benchmarks, but rather to stealthily embed a hidden backdoor. 
This backdoor would cause the model to produce a specific, malicious output when activated by a secret trigger phrase, while leaving its general capabilities, such as its MMLU score, largely intact. A high-performing, non-attacked 72B model typically achieves an MMLU score in the range of 85\% to 87\%; a successful stealth attack would not be expected to cause a major deviation from this range.
We conducted the experiment over 200 communication rounds, comparing the performance of the FLock-Protected system against three widely used but unprotected federated learning algorithms: FedAvg~\cite{mcmahan2017communication}, SCAFFOLD~\cite{karimireddy2020scaffold}, and FedAdam~\cite{reddiadaptive}. The results are visualized in Figure~\ref{fig:attack}.

\textbf{Analysis of Vulnerable Systems}. The first three plots of Figure~\ref{fig:attack} illustrate a consistent and critical security failure. In all three unprotected scenarios, the Backdoor ASR (Attack Success Rate), shown by the red dotted line, rises precipitously, saturating near 90\% within the first 40 communication rounds. This demonstrates that the attacker's backdoor was successfully and rapidly injected into the global model. Concurrently, the Performance (MMLU Score), shown in blue, stagnates around the 85.8\% baseline, failing to achieve the synergistic gains expected from collaborative learning. This outcome confirms that even advanced FL optimizers offer no inherent defense against model poisoning, resulting in a compromised and untrustworthy model.

The bottom-right plot tells a story of complete resilience. The Backdoor ASR remains flat at nearly zero for all $200$ rounds. This is direct empirical evidence that FLock’s stake-weighted validator committee successfully identified and rejected the malicious model updates submitted by the attacker, preventing the poison from corrupting the global model. Crucially, while the attack was being neutralized, the MMLU Score continued to improve steadily, climbing from the baseline to over 86.6\%.

In conclusion, this experiment provides definitive evidence that the FLock system's integrated security mechanisms are not only effective but essential for building trustworthy AI in decentralized environments.

\section{Conclusions}
This work confronts the critical challenges of security, scalability, and efficiency that have impeded decentralized large-scale language model fine-tuning. We provide the first empirical validation of successfully fine-tuning a 70B-parameter LLM within a secure, multi-domain, and fully decentralized framework, demonstrating that historical barriers are not insurmountable.

Our solution, the FLock system, integrates a blockchain-based trust layer yields unequivocal results: collaborative training leads to substantial gains in general capabilities and, critically, a more than 68\% reduction in attack success rates, showcasing superior adversarial robustness. Furthermore, the unified FLock model demonstrates synergistic knowledge transfer, achieving superior cross-domain generalization that often surpasses models trained in isolation on their own specialized data. This outcome refutes the notion that decentralized learning must compromise on performance, highlighting its potential to create more capable and resilient models.

Our architecture's resilience was definitively proven through its successful defense against a sophisticated backdoor poisoning attack, which readily compromised standard optimizers like FedAvg and SCAFFOLD. By identifying and rejecting malicious updates, FLock maintained model integrity while continuing to improve performance.

In conclusion, FLock presents a viable pathway toward a more democratized, secure, and cooperative AI ecosystem, enabling diverse organizations to build state-of-the-art models without relying on centralized entities. Future work will explore extending this framework to other modalities and refining its economic incentive structures.

\section*{Acknowledment}
We gratefully acknowledge the HongKong Generative AI
Research \& Development Center (HKGAI) for providing the substantial computational resources that were essential for this research. The experiments, particularly the large-scale fine-tuning of the 70B-parameter language model, were conducted on a high-performance computing cluster made available by HKGAI.
This powerful infrastructure, comprising 64 NVIDIA H800 GPUs, was indispensable for the successful execution of this work. We extend our sincere gratitude for their generous support. We would also like to thank Prof. Sirui Han from HKUST for coordinating the computational resources at HKGAI and Dr. Zhao Zhang from Rutgers University for the helpful discussions.

\bibliographystyle{icml2025}
\bibliography{refer}

\begin{thebibliography}{30}
\providecommand{\natexlab}[1]{#1}
\providecommand{\url}[1]{\texttt{#1}}
\expandafter\ifx\csname urlstyle\endcsname\relax
  \providecommand{\doi}[1]{doi: #1}\else
  \providecommand{\doi}{doi: \begingroup \urlstyle{rm}\Url}\fi

\bibitem[Bhagoji et~al.(2019)Bhagoji, Chakraborty, Mittal, and Calo]{bhagoji2019analyzing}
Bhagoji, A.~N., Chakraborty, S., Mittal, P., and Calo, S.
\newblock Analyzing federated learning through an adversarial lens.
\newblock In \emph{International conference on machine learning}, pp.\  634--643. PMLR, 2019.

\bibitem[Bharti(2023)]{gbharti2023financealpaca}
Bharti, G.
\newblock Finance alpaca.
\newblock \url{https://huggingface.co/datasets/gbharti/finance-alpaca}, 2023.

\bibitem[Che et~al.(2023)Che, Liu, Zhou, Ren, Zhou, Sheng, Dai, and Dou]{CheL00ZSDD23}
Che, T., Liu, J., Zhou, Y., Ren, J., Zhou, J., Sheng, V.~S., Dai, H., and Dou, D.
\newblock Federated learning of large language models with parameter-efficient prompt tuning and adaptive optimization.
\newblock In Bouamor, H., Pino, J., and Bali, K. (eds.), \emph{Proceedings of the 2023 Conference on Empirical Methods in Natural Language Processing, {EMNLP} 2023, Singapore, December 6-10, 2023}, pp.\  7871--7888. Association for Computational Linguistics, 2023.
\newblock \doi{10.18653/V1/2023.EMNLP-MAIN.488}.
\newblock URL \url{https://doi.org/10.18653/v1/2023.emnlp-main.488}.

\bibitem[Chen et~al.(2021)Chen, Tworek, Jun, Yuan, Pinto, Kaplan, Edwards, Burda, Joseph, Brockman, et~al.]{chen2021evaluating}
Chen, M., Tworek, J., Jun, H., Yuan, Q., Pinto, H. P. D.~O., Kaplan, J., Edwards, H., Burda, Y., Joseph, N., Brockman, G., et~al.
\newblock Evaluating large language models trained on code.
\newblock \emph{arXiv preprint arXiv:2107.03374}, 2021.

\bibitem[Dettmers et~al.(2023)Dettmers, Pagnoni, Holtzman, and Zettlemoyer]{dettmers2023qlora0}
Dettmers, T., Pagnoni, A., Holtzman, A., and Zettlemoyer, L.
\newblock Qlora: Efficient finetuning of quantized llms.
\newblock \emph{arXiv preprint arXiv: 2305.14314}, 2023.

\bibitem[Dong et~al.(2024)Dong, Wang, Sun, Kampffmeyer, Knottenbelt, and Xing]{dong2024defending}
Dong, N., Wang, Z., Sun, J., Kampffmeyer, M., Knottenbelt, W., and Xing, E.
\newblock Defending against poisoning attacks in federated learning with blockchain.
\newblock \emph{IEEE Transactions on Artificial Intelligence}, 2024.

\bibitem[Foo et~al.(2025)Foo, Prasad, and Khoo]{foo2025know}
Foo, J., Prasad, P.~S., and Khoo, S.
\newblock Know or not: a library for evaluating out-of-knowledge base robustness.
\newblock \emph{arXiv preprint arXiv:2505.13545}, 2025.

\bibitem[Geiping et~al.(2020)Geiping, Bauermeister, Dr{\"o}ge, and Moeller]{geiping2020inverting}
Geiping, J., Bauermeister, H., Dr{\"o}ge, H., and Moeller, M.
\newblock Inverting gradients-how easy is it to break privacy in federated learning?
\newblock \emph{Advances in neural information processing systems}, 33:\penalty0 16937--16947, 2020.

\bibitem[Guha et~al.(2023)Guha, Nyarko, Ho, Ré, Chilton, Narayana, Chohlas-Wood, Peters, Waldon, Rockmore, Zambrano, Talisman, Hoque, Surani, Fagan, Sarfaty, Dickinson, Porat, Hegland, Wu, Nudell, Niklaus, Nay, Choi, Tobia, Hagan, Ma, Livermore, Rasumov-Rahe, Holzenberger, Kolt, Henderson, Rehaag, Goel, Gao, Williams, Gandhi, Zur, Iyer, and Li]{guha2023legalbench}
Guha, N., Nyarko, J., Ho, D.~E., Ré, C., Chilton, A., Narayana, A., Chohlas-Wood, A., Peters, A., Waldon, B., Rockmore, D.~N., Zambrano, D., Talisman, D., Hoque, E., Surani, F., Fagan, F., Sarfaty, G., Dickinson, G.~M., Porat, H., Hegland, J., Wu, J., Nudell, J., Niklaus, J., Nay, J., Choi, J.~H., Tobia, K., Hagan, M., Ma, M., Livermore, M., Rasumov-Rahe, N., Holzenberger, N., Kolt, N., Henderson, P., Rehaag, S., Goel, S., Gao, S., Williams, S., Gandhi, S., Zur, T., Iyer, V., and Li, Z.
\newblock Legalbench: A collaboratively built benchmark for measuring legal reasoning in large language models, 2023.

\bibitem[Hendrycks et~al.(2021)Hendrycks, Burns, Basart, Zou, Mazeika, Song, and Steinhardt]{dan2021MMLU}
Hendrycks, D., Burns, C., Basart, S., Zou, A., Mazeika, M., Song, D., and Steinhardt, J.
\newblock Measuring massive multitask language understanding.
\newblock In \emph{ICLR 2021, Virtual Event, Austria, May 3-7, 2021}. OpenReview.net, 2021.
\newblock URL \url{https://openreview.net/forum?id=d7KBjmI3GmQ}.

\bibitem[Hu et~al.(2022)Hu, Shen, Wallis, Allen-Zhu, Li, Wang, Wang, Chen, et~al.]{hu2022lora}
Hu, E.~J., Shen, Y., Wallis, P., Allen-Zhu, Z., Li, Y., Wang, S., Wang, L., Chen, W., et~al.
\newblock Lora: Low-rank adaptation of large language models.
\newblock \emph{ICLR}, 1\penalty0 (2):\penalty0 3, 2022.

\bibitem[Hu et~al.(2023)Hu, Ganter, Deilamsalehy, Dernoncourt, Foroosh, and Liu]{huetal2023meetingbank}
Hu, Y., Ganter, T., Deilamsalehy, H., Dernoncourt, F., Foroosh, H., and Liu, F.
\newblock Meetingbank: A benchmark dataset for meeting summarization.
\newblock In \emph{Proceedings of the 61st Annual Meeting of the Association for Computational Linguistics (ACL)}, Toronto, Canada, 2023. Association for Computational Linguistics.

\bibitem[Jin et~al.(2019)Jin, Dhingra, Liu, Cohen, and Lu]{jin2019pubmedqa}
Jin, Q., Dhingra, B., Liu, Z., Cohen, W.~W., and Lu, X.
\newblock Pub{M}edqa: A dataset for biomedical research question answering.
\newblock \emph{arXiv preprint arXiv:1909.06146}, 2019.

\bibitem[Kairouz et~al.(2021)Kairouz, McMahan, Avent, Bellet, Bennis, Bhagoji, Bonawitz, Charles, Cormode, Cummings, et~al.]{kairouz2021advances}
Kairouz, P., McMahan, H.~B., Avent, B., Bellet, A., Bennis, M., Bhagoji, A.~N., Bonawitz, K., Charles, Z., Cormode, G., Cummings, R., et~al.
\newblock Advances and open problems in federated learning.
\newblock \emph{Foundations and trends{\textregistered} in machine learning}, 14\penalty0 (1--2):\penalty0 1--210, 2021.

\bibitem[Karimireddy et~al.(2020)Karimireddy, Kale, Mohri, Reddi, Stich, and Suresh]{karimireddy2020scaffold}
Karimireddy, S.~P., Kale, S., Mohri, M., Reddi, S., Stich, S., and Suresh, A.~T.
\newblock Scaffold: Stochastic controlled averaging for federated learning.
\newblock In \emph{International conference on machine learning}, pp.\  5132--5143. PMLR, 2020.

\bibitem[Kone{\v{c}}n{\`y} et~al.(2016)Kone{\v{c}}n{\`y}, McMahan, Yu, Richt{\'a}rik, Suresh, and Bacon]{konevcny2016federated}
Kone{\v{c}}n{\`y}, J., McMahan, H.~B., Yu, F.~X., Richt{\'a}rik, P., Suresh, A.~T., and Bacon, D.
\newblock Federated learning: Strategies for improving communication efficiency.
\newblock \emph{arXiv preprint arXiv:1610.05492}, 2016.

\bibitem[Li et~al.(2020)Li, Sahu, Talwalkar, and Smith]{li2020federated}
Li, T., Sahu, A.~K., Talwalkar, A., and Smith, V.
\newblock Federated learning: Challenges, methods, and future directions.
\newblock \emph{IEEE signal processing magazine}, 37\penalty0 (3):\penalty0 50--60, 2020.

\bibitem[Loukas et~al.(2021)Loukas, Fergadiotis, Androutsopoulos, and Malakasiotis]{loukas2021edgar}
Loukas, L., Fergadiotis, M., Androutsopoulos, I., and Malakasiotis, P.
\newblock {EDGAR}-{CORPUS}: Billions of tokens make the world go round.
\newblock In \emph{Proceedings of the Third Workshop on Economics and Natural Language Processing}, pp.\  13--18, Punta Cana, Dominican Republic, November 2021. Association for Computational Linguistics.
\newblock URL \url{https://aclanthology.org/2021.econlp-1.2}.

\bibitem[Mangrulkar et~al.(2022)Mangrulkar, Gugger, Debut, Belkada, Paul, and Bossan]{peft}
Mangrulkar, S., Gugger, S., Debut, L., Belkada, Y., Paul, S., and Bossan, B.
\newblock Peft: State-of-the-art parameter-efficient fine-tuning methods.
\newblock \url{https://github.com/huggingface/peft}, 2022.

\bibitem[McMahan et~al.(2017)McMahan, Moore, Ramage, Hampson, and y~Arcas]{mcmahan2017communication}
McMahan, B., Moore, E., Ramage, D., Hampson, S., and y~Arcas, B.~A.
\newblock Communication-efficient learning of deep networks from decentralized data.
\newblock In \emph{Artificial intelligence and statistics}, pp.\  1273--1282. PMLR, 2017.

\bibitem[PolicyAI(2024)]{PolicyAI2024NEPATEC}
PolicyAI.
\newblock Nepatec1.0: National environmental policy act text corpus 1.0.
\newblock \url{https://huggingface.co/datasets/PolicyAI/NEPATEC1.0}, 2024.

\bibitem[Rasley et~al.(2020)Rasley, Rajbhandari, Ruwase, and He]{rasley2020deepspeed}
Rasley, J., Rajbhandari, S., Ruwase, O., and He, Y.
\newblock Deepspeed: System optimizations enable training deep learning models with over 100 billion parameters.
\newblock In \emph{Proceedings of the 26th ACM SIGKDD international conference on knowledge discovery \& data mining}, pp.\  3505--3506, 2020.

\bibitem[Reddi et~al.(2021)Reddi, Charles, Zaheer, Garrett, Rush, Kone{\v{c}}n{\`y}, Kumar, and McMahan]{reddiadaptive}
Reddi, S.~J., Charles, Z., Zaheer, M., Garrett, Z., Rush, K., Kone{\v{c}}n{\`y}, J., Kumar, S., and McMahan, H.~B.
\newblock Adaptive federated optimization.
\newblock In \emph{International Conference on Learning Representations}, 2021.

\bibitem[Team(2024)]{qwen25}
Team, Q.
\newblock Qwen2.5: A party of foundation models, September 2024.
\newblock URL \url{https://qwenlm.github.io/blog/qwen2.5/}.

\bibitem[{\"U}st{\"u}n et~al.(2024){\"U}st{\"u}n, Aryabumi, Yong, Ko, D'souza, Onilude, Bhandari, Singh, Ooi, Kayid, et~al.]{ustun2024aya}
{\"U}st{\"u}n, A., Aryabumi, V., Yong, Z.-X., Ko, W.-Y., D'souza, D., Onilude, G., Bhandari, N., Singh, S., Ooi, H.-L., Kayid, A., et~al.
\newblock Aya model: An instruction finetuned open-access multilingual language model.
\newblock \emph{arXiv preprint arXiv:2402.07827}, 2024.

\bibitem[Wangyue~Li(2024)]{ai2challenge2024}
Wangyue~Li, Liangzhi~Li, T. X. X. L. W. D. N.~G.
\newblock Can multiple-choice questions really be useful in detecting the abilities of llms?
\newblock \emph{arXiv preprint arXiv:2403.17752}, 2024.

\bibitem[Xing et~al.(2021)Xing, Simeone, and Bi]{xing2021federated}
Xing, H., Simeone, O., and Bi, S.
\newblock Federated learning over wireless device-to-device networks: Algorithms and convergence analysis.
\newblock \emph{IEEE Journal on Selected Areas in Communications}, 39\penalty0 (12):\penalty0 3723--3741, 2021.

\bibitem[Zheng et~al.(2023)Zheng, Chiang, Sheng, Zhuang, Wu, Zhuang, Lin, Li, Li, Xing, et~al.]{zheng2023judging}
Zheng, L., Chiang, W.-L., Sheng, Y., Zhuang, S., Wu, Z., Zhuang, Y., Lin, Z., Li, Z., Li, D., Xing, E., et~al.
\newblock Judging llm-as-a-judge with mt-bench and chatbot arena.
\newblock \emph{Advances in Neural Information Processing Systems}, 36:\penalty0 46595--46623, 2023.

\bibitem[Zhuang et~al.(2023)Zhuang, Chen, and Lyu]{zhuang2023foundation}
Zhuang, W., Chen, C., and Lyu, L.
\newblock When foundation model meets federated learning: Motivations, challenges, and future directions.
\newblock \emph{arXiv preprint arXiv:2306.15546}, 2023.

\bibitem[Zou et~al.(2023)Zou, Wang, Carlini, Nasr, Kolter, and Fredrikson]{zou2023universal}
Zou, A., Wang, Z., Carlini, N., Nasr, M., Kolter, J.~Z., and Fredrikson, M.
\newblock Universal and transferable adversarial attacks on aligned language models.
\newblock \emph{arXiv preprint arXiv:2307.15043}, 2023.

\end{thebibliography}
\appendix
\section{LLM as a Judge}\label{appendix:llm-as-judge}
\begin{figure}[h]
    \centering
    \includegraphics[width=\linewidth]{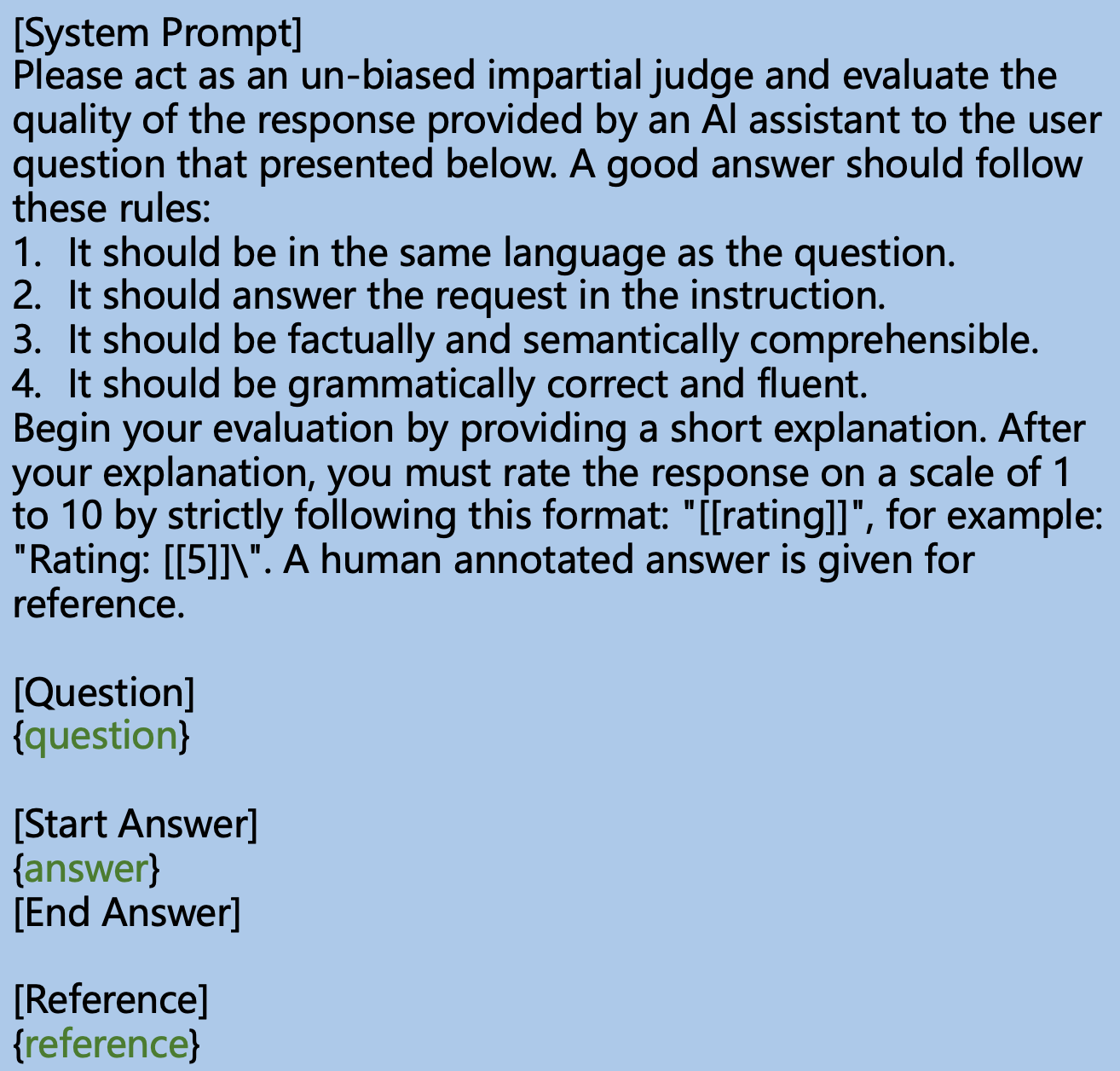}
    \caption{Prompt template used in GPT-4 Judge.}
    \label{fig:llm-as-a-judge}
\end{figure}

Our evaluation protocol leverages the GPT-4 Judge, utilizing the specific prompt template detailed in Figure~\ref{fig:llm-as-a-judge}. For the Aya dataset, we curated a test set from the original data, with each instance containing a distinct question and its associated ground-truth reference answer. The model being evaluated was prompted with each question to produce a candidate answer. Following this inference step, the question, the model's generated answer, and the reference answer were programmatically inserted into the respective ``question'', ``answer'', and ``reference'' fields of the template to facilitate a structured assessment.

\section{Hyperparameters of Experiments}


we fine-tuned \textbf{Qwen-2.5 72b} on 8 domain-specific datasets across 8 clients using \textbf{QLoRA}~\cite{dettmers2023qlora0} combined with \textbf{PEFT}~\cite{peft}. the fine-tuning employed 8-bit quantization, a learning rate of $3\times10^{-4}$, and a batch size of 16. we set the \emph{LoRA rank} to 16 with a scaling factor $\alpha=32$, targeting the \texttt{"k\_proj"} and \texttt{"v\_proj"} modules with a dropout rate of 0.05. Training was performed with a block size and cutoff length of 512. We used a warmup step of 1, a weight decay of 0.05, and conducted training over 3 local epochs and 20 local steps per client with 10 global communication rounds for FL.

\end{document}